\documentclass[11pt]{article}
\usepackage[margin=1in]{geometry}
\usepackage[utf8]{inputenc}
\usepackage[T1]{fontenc}
\usepackage[hidelinks]{hyperref}
\usepackage{mathtools}
\usepackage{amsmath,amsthm,amssymb,amsfonts,mathrsfs}
\usepackage{dsfont,dirtytalk,adjustbox,url,nicefrac,booktabs,array,todonotes,longtable,tabu,todonotes,multirow,tabularx,caption,RLDM,pifont,enumitem}
\usepackage{tikz-cd}
\usepackage{graphicx}
\usepackage{epsfig}
\usepackage{wrapfig}
\usepackage{comment}
\usepackage{bm}
\usepackage[english]{babel}
\usepackage[capitalise]{cleveref}
\usepackage{cite}
\newcommand{\cmark}{\ding{51}}%
\newcommand{\xmark}{\ding{55}}

\let\OLDthebibliography\thebibliography
\renewcommand\thebibliography[1]{
  \OLDthebibliography{#1}
  \setlength{\parskip}{0pt}
  \setlength{\itemsep}{0pt plus 0.3ex}
}

\newcommand{\E}{\mathbb E}

\newcommand{\defn}{\equiv}  

\newcommand{\kl}{\operatorname{D_{K L}}}
\newcommand{\ent}{\operatorname{H}}

\DeclareMathOperator*{\argmax}{arg\,max}

\theoremstyle{plain}
\newtheorem{theorem}{Theorem}[section]
\newtheorem{lemma}[theorem]{Lemma}

\theoremstyle{definition}
\newtheorem{definition}[theorem]{Definition}

\theoremstyle{remark}
\newtheorem{remark}[theorem]{Remark}

\title{Active Inference as a Model of Agency}
\author{%
  Lancelot Da Costa \\
  Department of Mathematics\\
  Imperial College London\\
  \texttt{l.da-costa@imperial.ac.uk} \\
  \And
  Samuel Tenka \\
  CSAIL \\
  Massachusetts Institute of Technology\\
  \texttt{c\,o\,l\,i{\tiny @}m\,i\,t.edu} \\
  \AND
  Dominic Zhao \\
  Common Sense Machines \\
  \texttt{dominic.zhao@csm.ai}\\
  \And
  Noor Sajid \\
  Wellcome Centre for Human Neuroimaging \\
  University College London \\
  \texttt{noor.sajid.18@ucl.ac.uk} \\
}

\begin{document}

\maketitle

\begin{abstract}
Is there a canonical way to think of agency beyond reward maximisation? In this paper, we show that any type of behaviour complying with physically sound assumptions about how macroscopic biological agents interact with the world canonically integrates exploration and exploitation in the sense of minimising risk and ambiguity about states of the world. This description, known as active inference, refines the free energy principle, a popular descriptive framework for action and perception originating in neuroscience. Active inference provides a normative Bayesian framework to simulate and model agency that is widely used in behavioural neuroscience, reinforcement learning (RL) and robotics. The usefulness of active inference for RL is three-fold. \emph{a}) Active inference provides a principled solution to the exploration-exploitation dilemma that usefully simulates biological agency. 
\emph{b}) It provides an explainable recipe to simulate behaviour, whence behaviour follows as an explainable mixture of exploration and exploitation under a generative world model, and all differences in behaviour are explicit in differences in world model.
\emph{c}) This framework is universal in the sense that it is theoretically possible to rewrite any RL algorithm conforming to the descriptive assumptions of active inference as an active inference algorithm. Thus, active inference can be used as a tool to uncover and compare the commitments and assumptions of more specific models of agency. 
\end{abstract}

\textbf{Keywords:} exploration, exploitation, expected free energy, generative world model, Bayesian inference.

\textbf{Acknowledgements}

The authors are indebted to Alessandro Barp, Guilherme França, Karl Friston, Mark Girolami, Michael I. Jordan and Grigorios A. Pavliotis for helpful input on a preliminary version to this manuscript. The authors thank Joshua B. Tenenbaum and MIT's Computational Cognitive Science group for interesting discussions that lead to some of the points discussed in this paper. LD is supported by the Fonds National de la Recherche, Luxembourg (Project code: 13568875) and a G-Research grant. This publication is based on work partially supported by the EPSRC Centre for Doctoral Training in Mathematics of Random Systems: Analysis, Modelling and Simulation (EP/S023925/1). NS is funded by the Medical Research Council (MR/S502522/1) and 2021–2022 Microsoft PhD Fellowship.

\pagebreak


\section{Introduction}

Reinforcement learning (RL) is a collection of methods that describe and simulate agency---how to map situations to actions---traditionally framed as optimising a numerical reward signal~\cite{bartoReinforcementLearningIntroduction1992}. The idea of maximising reward as underpinning agency is ubiquitous: with roots in utilitarianism~\cite{benthamIntroductionPrinciplesMorals2007} and expected utility theory~\cite{vonneumannTheoryGamesEconomic1944}, it also underwrites game theory~\cite{vonneumannTheoryGamesEconomic1944}, statistical decision-theory~\cite{bergerStatisticalDecisionTheory1985}, optimal control theory~\cite{bellmanDynamicProgramming1957, astromOptimalControlMarkov1965a}, and much of modern economics. From its inception, RL practitioners have supplemented reward seeking algorithms with various heuristics or biases geared towards simulating intelligent behaviour. Especially effective are intrinsic motivation or curiosity-driven rewards that 
encourage exploration~\cite{schmidhuber1991possibility,pathak2017curiosity,bellemare2016unifying}.
Thus, we ask: \emph{is there a canonical way to think of agency beyond reward maximisation?}

In this paper, we show that any behaviour complying with physically sound assumptions about how macroscopic biological agents interact with the world canonically integrates exploration and exploitation by minimising risk and ambiguity about external states of the world. This description, known as active inference, refines the free energy principle, a popular descriptive framework for action and perception birthed in neuroscience~\cite{fristonFreeEnergyPrinciple2022,fristonFreeenergyPrincipleUnified2010,fristonFreeEnergyPrinciple2006}.


Active inference provides a generic framework to simulate and model agency that is widely used in neuroscience~\cite{parrComputationalNeurologyMovement2021,pezzuloHierarchicalActiveInference2018,schwartenbeckComputationalMechanismsCuriosity2019,smithBayesianComputationalModel2020,isomuraCanonicalNeuralNetworks2022}, RL~\cite{mazzagliaContrastiveActiveInference2021,fountasDeepActiveInference2020,markovicEmpiricalEvaluationActive2021a,sajid2021exploration} and robotics~\cite{lanillosActiveInferenceRobotics2021,dacostaHowActiveInference2022a,catalRobotNavigationHierarchical2021,chameCognitiveMotorCompliance2020a}.
The usefulness of active inference for RL is three-fold. 
\emph{a}) Active inference provides an effective solution to the exploration-exploration dilemma that englobes the principles of expected utility theory~\cite{vonneumannTheoryGamesEconomic1944} and Bayesian experimental design~\cite{lindleyMeasureInformationProvided1956} and finesses the need for ad-hoc exploration bonuses in the reward function or decision-making objective.
\emph{b}) Active inference provides a transparent recipe to simulate behaviour by minimising risk and ambiguity with respect to an explicit generative world model. This enables safe and explainable decision-making by specifically encoding the commitments and goals of the agent in the world model~\cite{dacostaHowActiveInference2022a}.
\emph{c}) Active inference is universal in the sense that it is theoretically possible to rewrite any RL algorithm conforming to the descriptive assumptions of active inference as an active inference algorithm, e.g.,~\cite{baltieriPIDControlProcess2019}. Thus, active inference can be used as a tool to uncover and compare the commitments and assumptions of more specific models of agency.

\section{Deriving agency from physics}
\label{sec: deriving agency}
We describe systems that comprise an agent interacting with its environment. We assume that an agent and its environment
evolve together according to a stochastic process $x$. This definition entails a notion of time $\mathcal T$, which may be discrete or continuous, and a state space $\mathcal X$, which should be a measure space (e.g., discrete space, manifold, etc.). Recall that a stochastic process $x$ is a time-indexed collection of random variables $x_t$ on state space $\mathcal X$. Equivalently, $x$ is a random variable over trajectories on the state space $\mathcal T \to \mathcal X$.
We denote by $P$ the probability density of $x$ on the space of trajectories $\mathcal T \to \mathcal X$ (with respect to an implicit base measure).



In more detail, we factorise the state space $\mathcal X$ into states that belong to the agent $\mathcal H$ and states \emph{external} to the agent $\mathcal S$ that belong to the environment. Furthermore, we factorise agent's states into \emph{autonomous} states $\mathcal A$ and \emph{observable} states $\mathcal O$, respectively defined as the states which the agent does and does not have agency over.
In summary, the system $x$ is formed of \emph{external} $s$ and \emph{agent} processes $h$, the latter which is formed of \emph{observable} $o$, and \emph{autonomous} $a$ processes
$ \mathcal X \defn \mathcal S \times \mathcal H \defn \mathcal S \times \mathcal O \times \mathcal A \implies   x \defn (s, h) \defn (s, o, a)$.

\begin{figure}[t!]
\centering
    \includegraphics[width=\textwidth]{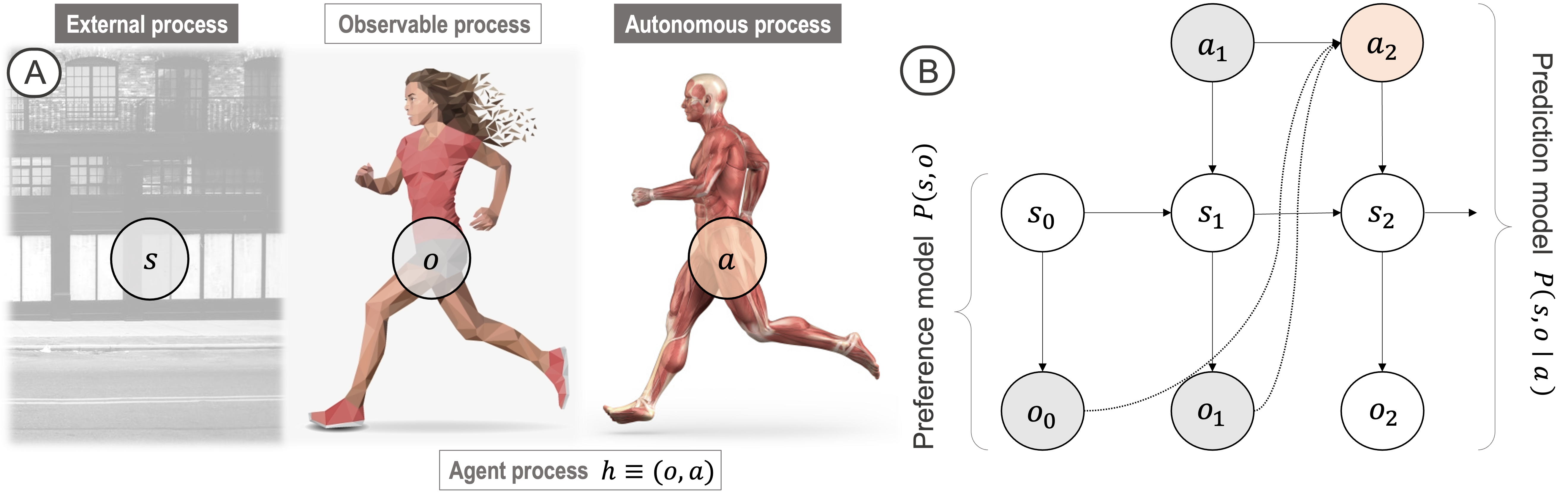}
    \caption{\textbf{(A)} This figure illustrates a human (agent process $h$) interacting with its environment (external process $s$), and the resulting partition into external $s$, observable $o$, and autonomous $a$ processes. The agent does not have direct access to the external process, but samples it through the observable process. The observable process constitutes the sensory epithelia (e.g., eyes and skin), which influences the environment through touch. The autonomous process constitutes the muscles and nervous system, which influences the sensory epithelia, e.g., by moving a limb, and the environment, e.g., through speech by activating vocal cords. Autonomous responses at time $t + \delta t$ may depend upon all the information available to the agent at time $t$, that is $h_{\leq t}$. Thus, the systems we are describing are typically \emph{non-Markovian}. \textbf{(B)} An active inference agent is completely described by its prediction model $P(s,o \mid a)$ and its preference model $P(s,o)$. When the prediction model is a POMDP the preference model is a hidden Markov model. In this setting, the colour scheme illustrates the problem of agency at $t=1$: the agent must execute an action (in red) based on previous actions and observations
    (in grey), which are informative about external states and future observations (in white). When specifying an active inference agent, it is important that prediction and preference models coincide on those parts they have in common; in this example the likelihood map $P(o \mid s)$. Note that these models need not be Markovian.} 
    \label{fig: partition}
\end{figure}

The description adopted so far could aptly describe particles interacting with a heat bath~\cite{pavliotisStochasticProcessesApplications2014,parrMarkovBlanketsInformation2020,rey-belletOpenClassicalSystems2006} as well as humans interacting with their environment (Figure~\ref{fig: partition}.A). We would like a description of macroscopic biological systems; so what distinguishes people from small particles? A clear distinction is that human behaviour occurs at the macroscopic level and, thus, is subject to classical mechanics. In other words, people are \emph{precise agents}:

\begin{definition}
\label{def: precise agent}
    An agent is \emph{precise} when it responds deterministically to 
    its environment, that is when $h \mid s$ is a deterministic process.
\end{definition}

\begin{remark}
\label{rem: coarse graining}
It is important not to conflate an agent's mental representations of external reality with external states; the former are usually an abstraction or coarse-grained representation of the latter. We do not consider agent's representations in this paper. \emph{We simply posit that there exists a detailed enough description of the environment that determines the agent's trajectory} (e.g., observations and actions). This will always be true for agents evolving according to classical mechanics.
\end{remark}

Agency means being in control one's actions and using them to influence the environment~\cite{haggard2009experience}.
At time $t$, the information available to the agent is its past trajectory $h_{\leq t}=(o_{\leq t},a_{\leq t})$, i.e. its history. We define \emph{decision-making} as a choice of autonomous trajectory in the future $a_{>t}$ given available knowledge $h_{\leq t}$. Furthermore, we define \emph{agency} as the process through which an agent makes and executes decisions. We interpret $P(s,o \mid h_{\leq t})$ as expressing the agent's \textit{preferences} over environmental and observable trajectories given available data, and $P(s,o \mid a_{>t},h_{\leq t})$ as expressing the agent's \textit{predictions} over environmental and observable paths given a decision (e.g., Figure~\ref{fig: partition}.B). Crucially, agency is governed by an \emph{expected free energy} functional of the agent's predictions and preferences (Appendix~\ref{app: derivations})
\begin{align}
\label{eq: EFE}
-\log P(a_{>t} \mid h_{\leq t}) = \E_{P(s, o \mid a_{>t}, h_{\leq t})}[\log P(s \mid  a_{>t}, h_{\leq t})- \log P(s,o \mid h_{\leq t})]. \tag{EFE}
\end{align}
We have formulated agency as optimising an objective: the lower the expected free energy, the more likely a course of action, and vice-versa.

\section{Characterising agency}
\label{sec: characterising agency}

This description of agency combines many accounts of behaviour that predominate in cognitive science and engineering. Indeed, decomposing the expected free energy~\eqref{eq: EFE} reveals fundamental imperatives that underwrite agency, such as minimising risk and ambiguity (Appendix~\ref{app: derivations}):
\begin{equation}
\label{eq: risk ambiguity}
\begin{split}
    -\log P(a_{>t} \mid h_{\leq t}) = \underbrace{\kl\big[\,\overbrace{P(s \mid  a_{>t}, h_{\leq t})}^{\text{predicted paths}} \mid \overbrace{P( s\mid h_{\leq t})}^{\text{preferred paths}}\,\big]}_{\text{risk}} +
    \underbrace{\E_{P(s \mid a_{>t}, h_{\leq t})}\big[\ent[P(o \mid s , h_{\leq t})]\big]}_{\text{ambiguity}}.
\end{split}
\end{equation}
\textbf{Risk} refers to the KL divergence between the predicted and preferred external course of events. 
As minimising a reverse KL divergence leads to mode matching behaviour~\cite{minkaDivergenceMeasuresMessage2005}, minimising risk leads to \emph{risk-averse} decisions that avoid disfavoured courses of events (c.f., Appendix~\ref{app: simulations}). 
In turn, risk-aversion is a hallmark of prospect theory, which describes human choices under discrete alternatives with no ambiguity~\cite{kahnemanProspectTheoryAnalysis1979}.
Additionally, risk is the main decision-making objective in modern approaches to control as inference~\cite{levineReinforcementLearningControl2018,rawlikStochasticOptimalControl2013,toussaintRobotTrajectoryOptimization2009,millidgeRelationshipActiveInference2020a}, variously known as Kalman duality~\cite{kalmanNewApproachLinear1960,todorovGeneralDualityOptimal2008a}, KL control~\cite{kappenOptimalControlGraphical2012} and maximum entropy RL~\cite{ziebartModelingPurposefulAdaptive2010}.

\textbf{Ambiguity} refers to the expected entropy of future observations, given future external trajectories. An external trajectory that may lead to various distinct observation trajectories is highly ambiguous---and vice-versa. Thus, minimising ambiguity leads to sampling observations that enable to recognise the external course of events. This leads to a type of observational bias commonly known as the \emph{streetlight effect} or \emph{drunkard's search} in behavioural psychology~\cite{kaplanConductInquiry1973}: when a person loses their keys at night, they initially search for them under the streetlight because the resulting observations ("I see my keys under the streetlight" or "I do not see my keys under the streetlight") accurately disambiguate external states of affairs.

Additionally, agency maximises extrinsic and intrinsic value (Appendix~\ref{app: derivations}):
\begin{align}
\label{eq: EU + BOD}
   -\log P(a_{>t} \mid h_{\leq t}) 
  &\geq -\underbrace{\mathbb{E}_{P(o \mid a_{>t} , h_{\leq t})}\big[\log \overbrace{P\left(o \mid h_{\leq t}\right)}^{\text{preferred paths}}\big]}_{\text{extrinsic value}} -\underbrace{\mathbb{E}_{P(o \mid a_{>t}, h_{\leq t})}\big[\kl\left[P\left(s \mid o, a_{>t}, h_{\leq t}\right) \mid P\left(s \mid a_{>t}, h_{\leq t}\right)\right]\big]}_{\text{intrinsic value}}.
\end{align}
\textbf{Extrinsic value} refers to the (log) likelihood of observations under the model of preferences. Maximising extrinsic value leads to sampling observations that are likely under the model of preferences. Through the correspondence between log probabilities and utility functions~\cite{kalmanNewApproachLinear1960,bridleProbabilisticInterpretationFeedforward1990,luceIndividualChoiceBehavior1959} this is equivalent to maximising expected utility or expected reward. This underwrites expected utility theory~\cite{vonneumannTheoryGamesEconomic1944}, game theory~\cite{vonneumannTheoryGamesEconomic1944}, optimal control~\cite{bellmanDynamicProgramming1957, astromOptimalControlMarkov1965a} and RL~\cite{bartoReinforcementLearningIntroduction1992}. Bayesian formulations of maximising expected utility under uncertainty are also known as Bayesian decision theory~\cite{bergerStatisticalDecisionTheory1985}.

\textbf{Intrinsic value} refers to the amount of information gained about external courses of events under a decision. 
Favouring decisions to maximise information gain leads to a goal-directed form of exploration~\cite{schwartenbeckComputationalMechanismsCuriosity2019}, driven to answer "what would happen if I did that?"~\cite{schmidhuberFormalTheoryCreativity2010}. Interestingly, this decision-making procedure underwrites Bayesian experimental design~\cite{lindleyMeasureInformationProvided1956} and active learning in statistics~\cite{mackayInformationBasedObjectiveFunctions1992}, intrinsic motivation and artificial curiosity in machine learning and robotics~\cite{oudeyerWhatIntrinsicMotivation2007,schmidhuberFormalTheoryCreativity2010,bartoNoveltySurprise2013,sunPlanningBeSurprised2011,deciIntrinsicMotivationSelfDetermination1985}. This is mathematically equivalent to optimising expected Bayesian surprise and mutual information, which underwrites visual search~\cite{ittiBayesianSurpriseAttracts2009,parrGenerativeModelsActive2021} and the organisation of our visual apparatus~\cite{barlowPossiblePrinciplesUnderlying1961,linskerPerceptualNeuralOrganization1990,opticanTemporalEncodingTwodimensional1987a}.

We have formulated agent's behaviour as minimising an objective that canonically weighs an exploitative term with an explorative term---risk plus ambiguity---which 
provides a principled solution to the exploration-exploitation dilemma~\cite{berger-talExplorationExploitationDilemmaMultidisciplinary2014}.

\section{Simulating agency}
\label{sec: simulating agency}

Active inference specifies an agent by a \textit{prediction model} $P(s,o \mid a)$, expressing the distribution over external and observable paths under autonomous paths, and a \textit{preference model} $P(s,o)$, expressing the preferred external and observable trajectories (e.g., Figure ~\ref{fig: partition}.B). In discrete time, agency proceeds by approximating the expected free energy given past observations and actions $h_{\leq t}=(o_{\leq t}, a_{\leq t})$ and use it to govern agency (details and simulations in Appendices~\ref{app: details algo} and~\ref{app: simulations}):
\begin{enumerate}
\item \emph{Preferential inference:} infer preferences about external and observable trajectories, i.e., approximate $P(s,o \mid h_{\leq t})$ with $Q(s,o \mid h_{\leq t})$ (e.g., Figure~\ref{fig: pref infer}).
\item For each possible sequence of 
future actions $a_{>t}$:
\begin{enumerate}
    \item \emph{Perceptual inference:} infer external and observable paths under the action sequence, i.e., approximate $P(s,o \mid~a_{>t} , h_{\leq t})$ with $Q(s,o \mid a_{>t}, h_{\leq t})$.
    \item \emph{Planning as inference:} assess the action sequence by evaluating its expected free energy~\eqref{eq: EFE}, i.e., 
    \begin{equation*}
    \label{eq: planning as inference}
        -\log Q(a_{>t} \mid h_{\leq t}) \defn \E_{Q(s, o \mid a_{>t}, h_{\leq t})}\big[\log Q(s \mid  a_{>t}, h_{\leq t})- \log Q(s,o \mid h_{\leq t})\big].
    \end{equation*}
\end{enumerate}
    \item \emph{Agency:} execute the most likely decision $\boldsymbol a_{t+1}$, or use $Q(a_{t+1} \mid h_{\leq t})$ as a model for behavioural data, where
    \begin{equation*}
        \boldsymbol a_{t+1} = \argmax Q(a_{t+1} \mid h_{\leq t}), \qquad
        Q(a_{t+1} \mid h_{\leq t})= \sum_{a_{>t}} Q(a_{t+1} \mid a_{>t}) Q(a_{>t} \mid h_{\leq t}). \label{eq: decision-making}
    \end{equation*}
\end{enumerate}

\section{Concluding remarks}
Active inference is a description of macroscopic biological agents derived from physics, which provides a generic framework to model biological and artificial behaviour. It provides a transparent description of agency in terms of minimising risk and ambiguity (i.e., an expected free energy functional) under some prediction and preference models about the world. 
This formulation is universal in the sense that any RL agent satisfying the descriptive assumptions of active inference (Appendix~\ref{app: derivations}) is behaviourally equivalent to an active inference agent under some implicit prediction and preference models.

Active inference provides an effective solution to the exploration-exploitation dilemma~\cite{berger-talExplorationExploitationDilemmaMultidisciplinary2014} that finesses the need for ad-hoc exploration bonuses in the reward function or decision-making objective (Appendices~\ref{app: simulations} and~\ref{app: comparing objectives})~\cite{ball2020ready,zintgraf2019varibad,hafner2019dream} . In particular, the expected free energy englobes various objectives used to describe or simulate behaviour across cognitive science and engineering, endowing it with various useful properties such as information-sensitivity (Appendix~\ref{app: comparing objectives}). Perhaps the closest RL formulations are action and perception as divergence minimisation~\cite{hafnerActionPerceptionDivergence2020}, which considers a similar decision-making objective; control as inference, which can be seen as minimising risk but not ambiguity~\cite{levineReinforcementLearningControl2018,millidgeRelationshipActiveInference2020a,ziebartModelingPurposefulAdaptive2010}; and Hyper~\cite{zintgraf2021exploration}, which proposes reward maximisation alongside minimising uncertainty over both external states and model parameters. 

Active inference provides a recipe for safe algorithmic decision-making. All behaviour under active inference is explainable as a mixture of exploration and exploitation under a generative world model, and all differences in behaviour are explicit in differences in world model. In particular, the agent's commitments and goals can be explicitly encoded in the world model by the user. In addition, the expected free energy leads to risk-averse decisions (Appendix~\ref{app: comparing objectives}), a crucial feature for engineering applications where catastrophic consequences are possible, e.g., robot-assisted surgery~\cite{dacostaHowActiveInference2022a}.

While active inference clarifies how agency unfolds given a generative world model representing how the environment causes observations (Section~\ref{sec: simulating agency}), it does not specify which representations underwrite highly intelligent behaviour. 
Promising steps in this direction include employing hierarchical probabilistic generative models with deep neural networks~\cite{soltani2017shapenet,fristonDeepTemporalModels2018,catalRobotNavigationHierarchical2021,parrComputationalNeurologyMovement2021,gothoskar2021threedpthree}. 
Beyond this, we conclude by asking: what kinds of generative models do humans use to represent their environment? And: what are the computational mechanisms under which a child’s mind develops into an adult mind by gradually learning its world model~\cite{tervoNeuralImplementationStructure2016,lakeBuildingMachinesThat2016,ullmanBayesianModelsConceptual2020,goodmanLearningTheoryCausality2011}?


\bibliographystyle{unsrt}
\bibliography{bib.bib}

\appendix

\section{Details on the active inference algorithm}
\label{app: details algo}

One specifies an active inference agent by a \textit{prediction model} $P(s,o \mid a)$ and a \textit{preference model} $P(s,o)$. Taking the example of Figure~\ref{fig: partition}.B, the prediction model may be a \textit{partially observable Markov decision process} (POMDP). A POMDP is a discrete time model of how actions influence external and observable trajectories. In a POMDP, 1) each external state depends only on the current action and previous external state $P(s_t \mid  s_{t-1}, a_t)$, and 2) each observation depends only on the current external state $P(o_t \mid s_t)$. If one additionally specifies 3) a distribution of preferences over external trajectories $P(s)$, one obtains a (hidden Markov) preference model by combining 2) \& 3). In general, the preference model may be specified independently from the prediction model, however, since these are the same distribution conditioned on different variables it is important that they coincide on the parts they have in common; in this example, the likelihood map $P(o_t \mid s_t)$. Importantly, the generative models specify the temporal horizon of the agent (i.e., how much time ahead it plans and represents the world).

Under this model specification, agency is governed by the expected free energy~\eqref{eq: EFE}
\begin{align*}
-\log P(a_{>t} \mid h_{\leq t}) = \E_{P(s, o \mid a_{>t}, h_{\leq t})}[\log P(s \mid  a_{>t}, h_{\leq t})- \log P(s,o \mid h_{\leq t})]. 
\end{align*}

\begin{wrapfigure}[34]{r}{0.5\textwidth}
    \centering
    \includegraphics[width=0.49\textwidth]{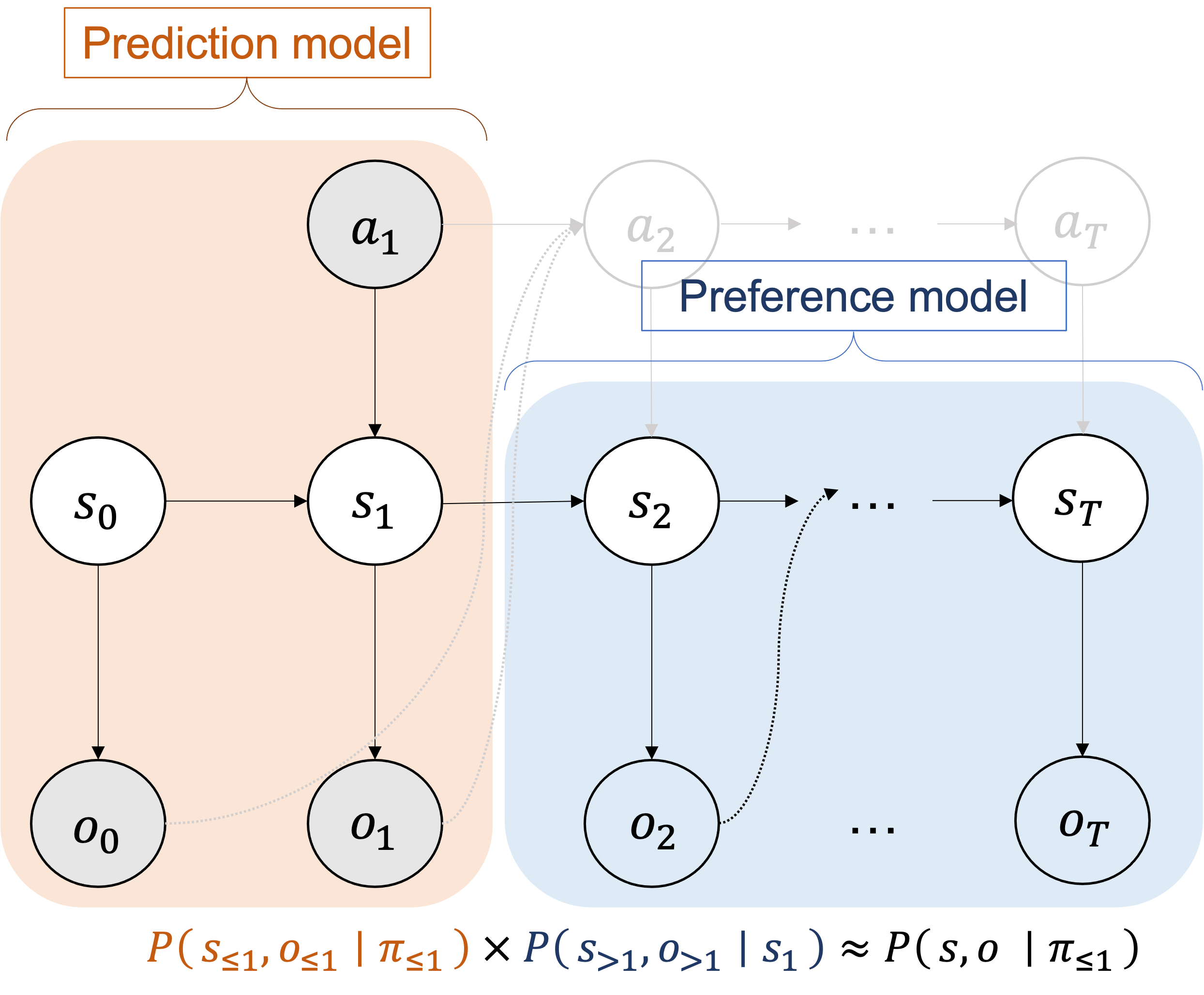}
    \caption{\textbf{Preferential inference on POMDPs}. This figure illustrates how one may infer preferences efficiently when the prediction model is a POMDP and the preference model is a hidden Markov model (c.f., Figure~\ref{fig: partition}.B). The problem is illustrated at time $t=1$. We may approximate the posterior preferences over states and observations given available data $P(s,o \mid h_{\leq t})$ by the product of the predictions of past states and observations given available data $P(s_{\leq t},o_{\leq t} \mid h_{\leq t})$ and the preferences over future states and observations given the current state $P(s_{> t},o_{> t} \mid s_{t})$, c.f.,~\eqref{eq: pref inference factorisation}. These two distributions can, in turn, be obtained from the prediction and preference models via approximate inference. More simply, when preferences $P(s_t,o_t)$ are i.i.d. for all $t$ (c.f., Figure~\ref{fig: T-Maze}), the latter distribution simply equals future preferences $P(s_{> t},o_{> t})$.}
    \label{fig: pref infer}
\end{wrapfigure}

The standard active inference algorithm (Section~\ref{sec: simulating agency}) first approximates the expected free energy by approximating the usually intractable posterior distributions within. For instance, \emph{preferential inference} involves inferring preferences given previous actions and observations $P(s,o \mid h_{\leq t})$. Observe that this distribution factorises
\begin{equation}
\label{eq: pref inference factorisation}
    P(s,o \mid h_{\leq t}) = P(s_{\leq t},o_{\leq t} \mid h_{\leq t})P(s_{> t},o_{> t} \mid x_{\leq t})
\end{equation}
leading to efficient approximation in a variety of cases (e.g., Figure~\ref{fig: pref infer}). More simply, \emph{perceptual inference} simply involves conditioning the prediction model on past observations
\begin{equation*}
    P(s,o \mid a_{>t}, h_{\leq t}) = P(s,o \mid a, o_{\leq t}).
\end{equation*}
Once one has obtained an approximate posterior distribution over the next action $Q(a_{t+1} \mid h_{\leq t})$, one can either execute the most likely action as shown in Section~\ref{sec: simulating agency}, or simulate generic behaviour by sampling actions from the posterior distribution $$a_{t+1} \sim Q(a_{t+1} \mid h_{\leq t})$$
(not shown in Section~\ref{sec: simulating agency}). The latter is to be employed when using active inference as a generative model for behavioural data.

There are many ways to scale the active inference algorithm for practical applications~\cite{barpGeometricMethodsSampling2022}. These generally involve employing hierarchical generative models with deep neural networks~\cite{fristonDeepTemporalModels2018,catalRobotNavigationHierarchical2021}, structured mean-field approximations~\cite{dacostaActiveInferenceDiscrete2020,heinsPymdpPythonLibrary2022,smithStepbyStepTutorialActive2022,parrActiveInferenceFree2022,schwobelActiveInferenceBelief2018}, amortisation~\cite{fountasDeepActiveInference2020}, and Monte-Carlo tree search~\cite{fountasDeepActiveInference2020,championBranchingTimeActive2021,championBranchingTimeActive2021a,maistoActiveTreeSearch2021}. Furthermore, this algorithm can also be used to learn the agent's prediction and preference models~\cite{barpGeometricMethodsSampling2022}.

\section{Simulations of behaviour}
\label{app: simulations}

\begin{figure}[t!]
\centering
    \includegraphics[width=0.6\textwidth]{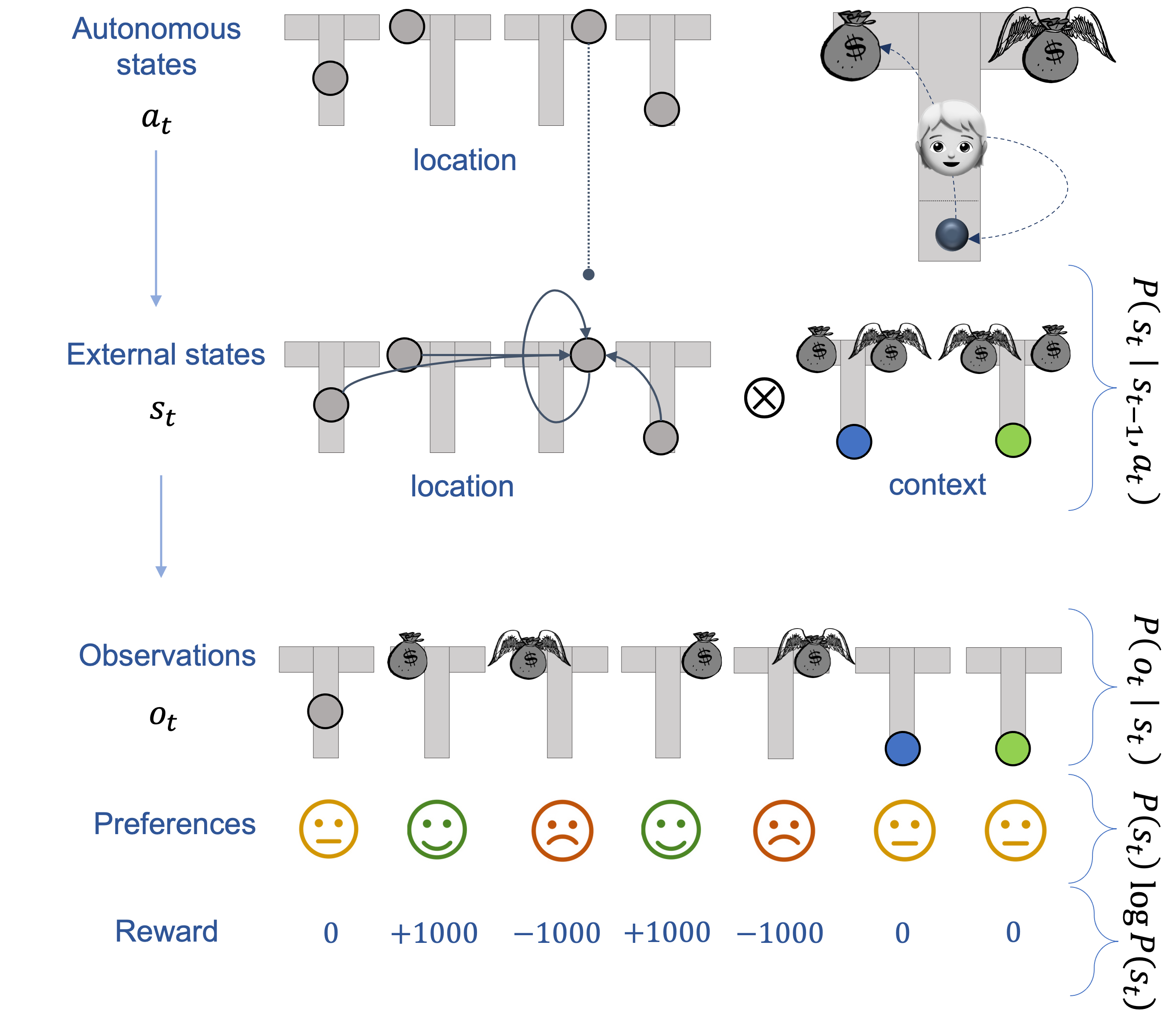}
    \caption{\textbf{T-Maze environment.} This Figure illustrates a simple sequential decision-making task with a temporal horizon of $2$. In other words, the agent must choose $a_1$ given $h_0=o_0$ and, subsequently, $a_2$ given $h_{\leq 1}= (a_1, o_{\leq 1})$. In more detail, $s_t$: The T-Maze has four possible spatial \emph{locations}: middle, top-left, top-right, bottom. One of the top locations yields a reward of 1000\$ (money bag), while the other yields a punishment of -1000\$ (flying money bag)---per time-step spent in respective locations. The reward's location determines the \emph{context}. The bottom arm contains a cue whose colour (blue or green) discloses the context. Together, location and context determine the external state. $o_t$: The agent always observes its spatial location. In addition, when it is at the top of the Maze, it receives the reward or the punishment; when it is at the bottom, it observes the colour of the cue. $a_t$: Each action corresponds to visiting one of the four spatial locations. $P(s_t)$: The agent prefers being at the reward's location ($-\log P(s_t) =1000$) and avoid the punishment's location ($-\log P(s_t) =-1000$). All other external states have a neutral preference ($-\log P(s_t)=0$). $s_0$: the agent starts in the middle location and the context is initialised at random. Together these datum determine prediction $P(s,o \mid a)$ and preference $P(s,o)$ models, which in turn, determine agency by minimisation of expected free energy (Section~\ref{sec: simulating agency}).} 
    \label{fig: T-Maze}
\end{figure}

Here is a simple simulation of behaviour in a T-Maze environment (Figure~\ref{fig: T-Maze}). The environment has four spatial locations (top left, top right, middle and bottom). One of the top arms yields a reward of $1000\$$ while the other yields a loss of $1000\$$---per time-step spent in each respective location. The two remaining locations have no reward ($0\$$). Suppose that you start in the middle arm while being unaware of the reward's location. The task has two time-steps, so you may visit two (not necessarily distinct) locations of the Maze in the order you wish. At first, you may: \emph{a}) stay where you are, \emph{b}) go to the bottom arm to collect a cue that discloses the reward's location, \emph{c}) go to one of the top arms to determine the reward's location by elimination, even though this risks receiving the punishment. Options \emph{b}) and \emph{c}) both enable you to collect $1000\$$ upon visiting the reward's location at the second step of the task. In choosing policy \emph{b}) the payoff is $1000\$$ with certainty. In choosing policy \emph{c}) the payoff is variable: $2000\$, 1000\$$ or $0\$$ with $1/3$ probability each. In choosing policy \emph{a}), the payoff is strictly worse. \emph{Which of these options a), b) or c) will you choose?}

In humans, the most common choice is option (\emph{b})---visiting the bottom arm to collect the cue, thereby inferring the reward's location---and subsequently collecting the reward~\cite{kahnemanProspectTheoryAnalysis1979}, while always avoiding the punishment. We now compare the behaviours of active inference agents (Section~\ref{sec: simulating agency}), reward maximising agents, and reward plus information gain maximising agents:

\begin{enumerate}[wide, labelwidth=!, labelindent=0.2pt]
    \item \textbf{Active inference agent.} $o_0$: The agent is in the middle of the Maze and is unaware of the context. $a_1$: Visiting the bottom or top arms have a lower ambiguity than staying, as they yield observations that disclose the context. However, staying or visiting the bottom arm are safer options, as visiting a top arm risks receiving the punishment. By acting to minimise both risk and ambiguity~\eqref{eq: risk ambiguity} the agent goes to the bottom. $o_1$: The agent observes the cue and hence determines the context. $a_2$: All actions have equal ambiguity as the context is known. Collecting the reward has a lower risk than staying or visiting the middle, which themselves have a lower risk than collecting the punishment. Thus, the agent visits the arm with the reward. See~\cite{fristonActiveInferenceProcess2017,schwartenbeckComputationalMechanismsCuriosity2019} for more details. In summary, the active inference agent ends the task with $1000\$$ with certain probability. \label{item: AIF agent}
    \item \textbf{Expected reward maximising agent.} $o_0$: idem. $a_1$: all locations have the same expected reward of $0\$$, so it is unclear where to go. Suppose then that the agent chooses an action at random according to an uniform distribution. $o_1$: In the event that it has chosen to leave the middle location (i.e., $75\%$ of the time), this enables it to unambiguously determine the reward's location. $a_2$: the agent collects the reward. In summary, $75\%$ of the time, the agent ends the task with $2000\$, 1000\$$ or $0\$$ with $1/3$ probability, respectively. $25\%$ of the time, it's performance will be worse: $1000\$, -1000\$$ with $1/4$ probability, respectively, and $0\$$ with probability $1/2$ probability. 
    \label{item: R agent}
    \item \textbf{Expected reward plus information gain maximising agent.} $o_0$: idem. $a_1$: all locations have the same expected reward of $0\$$ and the bottom and top locations give information as to the reward's location. Therefore, in maximising the sum of expected utility and expected information gain the agent leaves the middle location. $a_2$: the agent collects the reward. In summary, the agent ends the task with $2000\$, 1000\$$ or $0\$$ with $1/3$ probability, respectively. \label{item: RIG agent}
\end{enumerate}

We analyse these differences in behaviour in Appendix~\ref{app: comparing objectives}. For more complex simulations of sequential decision-making using active inference, please see~\cite{sajidActiveInferenceDemystified2021,fristonActiveInferenceCuriosity2017,parrComputationalNeurologyActive2019,fountasDeepActiveInference2020,millidgeDeepActiveInference2020,catalRobotNavigationHierarchical2021,fristonDeepTemporalModels2018}. 

\section{Comparing expected free energy, expected reward, and expected reward plus information gain}
\label{app: comparing objectives}

So what distinguishes these objectives? Following the simulations of Appendix~\ref{app: simulations} we distinguish two main features: \emph{information-sensitivity}, i.e., encoding the value of information, and \emph{risk-aversion} (see Table~\ref{table: comparison}). Briefly,

\begin{table}[h!]
\centering
\begin{tabular}{c c c}
 \hline
  Objective & Information-sensitive & Risk-averse  \\
 \hline
             {Expected free energy}	&\cmark  & \cmark \\
             {Expected reward}	& \xmark  & \xmark \\
             {Expected reward plus information gain}  & \cmark & \xmark \\
 \hline
\end{tabular}
\caption{Comparing objectives.} \label{table: comparison}
\end{table}

\begin{enumerate}[wide, labelwidth=!, labelindent=0.2pt]
    \item \textbf{Information-sensitivity.} Optimising expected reward does not intrinsically value the information gained following an action. In contrast, the expected free energy, or an expected information gain add-on to expected reward, both select actions that value the information afforded by an action.
    \item \textbf{Risk-aversion.} Expected reward and expected information-gain maximising agents cannot distinguish between two alternatives with the same expected reward and information gain. In that sense, they are perfectly rational agents. In contrast, active inference agents are risk-averse to the extent that they will avoid choices that can yield to a large loss, even when the expected payoff is neutral. Being risk-averse in this sense has two advantages: \emph{a}) in modelling human behaviour, as this is a defining feature of human decision-making~\cite{kahnemanProspectTheoryAnalysis1979}. \emph{b}) In engineering applications where the actions of artificial agents can potentially have catastrophic consequences---such as in algorithmic trading, human-robot interaction and robot-assisted surgery---risk-aversion enables safe decision-making by actively avoiding negative outcomes~\cite{dacostaHowActiveInference2022a}.
\end{enumerate}

\paragraph{Inducing information-sensitivity and risk-aversion by tuning the reward function.} Note that we can endow expected reward (ER) and expected reward plus information gain (ERIG) agents with information-sensitivity and risk-aversion by simply tuning the reward function. For instance, in the simulations of Appendix~\ref{app: simulations}, we can penalise the middle location of the T-Maze by $-1\$$ as it affords no information. In this case, the ER and ERIG maximising agents will behave the same as the initial ERIG maximising agent (Simulation~\ref{item: RIG agent}). Furthermore, in remarking that losing $1000\$$ hurts more than the gain afforded by receiving $1000\$$ we may update the (negative) reward associated with the loss to $-1001$. With this the ER and ERIG maximising agents will behave exactly as the active inference agent with the initial reward function (Simulation~\ref{item: AIF agent}).

Yet, tuning the reward function to achieve desired behaviour has two main drawbacks. For example, should we score the punishment by $-1001$ or $-1005$? There is no definite answer. Though these will not affect behaviour in the T-Maze task, this will not be the case in complex environments where different reward functions will lead to (sometimes unexpectedly) distinct behaviour. Secondly, it is well-known that manually tuning the reward to produce desired behaviour is a difficult and impractical task, especially in complex or changing environments.

We conclude that it is best to work with the ground truth reward whenever this is unambiguously defined (e.g., monetary reward or payoff by reaching a goal) and for information-sensitivity and risk-aversion to be incorporated in the agent's decision-making objective. Though many approaches to RL integrate information-sensitivity~\cite{zintgraf2021exploration,hafner2019dream} and risk-aversion~\cite{ziebartModelingPurposefulAdaptive2010,levineReinforcementLearningControl2018} with reward maximisation, active inference allows to canonically and transparently integrate these imperatives in the objective function.

\section{Assumptions and mathematical derivations}
\label{app: derivations}

This Appendix lists the different assumptions and mathematical arguments that underwrite the derivation of active inference presented in Sections~\ref{sec: deriving agency} and~\ref{sec: characterising agency}. 

\textbf{Assumptions:}
\begin{enumerate}
    \item \textbf{Stochastic process}: Agent and environment evolve according to a stochastic process $x\equiv (s,h)\equiv (s,o,a)$.
    \item \textbf{Precise agent}: The description of external state space is sufficiently detailed such that $h \mid s$ is a deterministic process. 
    \item \textbf{Countability of the path space}: The path space $\mathcal T \to \mathcal X$ is countable (e.g., time $\mathcal T$ is countable and the state space $\mathcal X$ is finite, or vice-versa). Presumably, our results can be extended to uncountable path spaces by a limiting argument. This is left for future work.
\end{enumerate}

Under the countability assumption we may take $P$ to be the probability density of the process w.r.t. the uniform measure. We have~\cite{fristonPathIntegralsParticular2022}:

\begin{equation*}
\begin{split}
    -\log P(a_{>t} \mid h_{\leq t}) &= \E_{P(s,o\mid a_{>t} ,h_{\leq t})}[-\log P(a_{>t} \mid h_{\leq t})]\\
    &= \E_{P(s,o\mid a_{>t} ,h_{\leq t})}[\log P(s,o \mid a_{>t},h_{\leq t})-\log P(s,o, a_{>t} \mid h_{\leq t})]\\
    &= \E_{P(s,o\mid a_{>t} ,h_{\leq t})}[\log P(o \mid s,a_{>t}, h_{\leq t})+ \log P(s \mid a_{>t}, h_{\leq t})-\log P(a_{>t} \mid s,o,h_{\leq t})- \log P(s,o  \mid h_{\leq t})]. 
\end{split}
\end{equation*}

\begin{lemma}
\label{lemma: normalisation}
Under the countability and precise agent assumptions, we have for any value of $a_{>t},h_{\leq t}$
\begin{align*}
    \E_{P(s,o\mid a_{>t} ,h_{\leq t})}[\log P(o \mid s,a_{>t}, h_{\leq t})-\log P(a_{>t} \mid s,o,h_{\leq t})]=0.
\end{align*}
\end{lemma}

\begin{proof}
By the precise agent assumption, $s$ determines $(o,a_{>t})$. In other words, there are functions $f: (\mathcal T \to \mathcal S) \to (\mathcal T \to \mathcal O)$ and $g: (\mathcal T \to \mathcal S) \to (\mathcal T_{>t} \to \mathcal A)$ such that $(o,a_{>t}) = (f ,g) \circ s$. In particular,
\begin{align*}
P(o \mid s, a_{>t}, h_{\leq t})&=P(o \mid s, o_{\leq t}) \propto P(o \mid s)= \delta_{f(s)}(o), \quad  P(a_{>t} \mid s,o, h_{\leq t})=P(a_{>t} \mid s)= \delta_{g(s)}(a_{>t}),\\
P(s,o \mid a_{>t}, h_{\leq t})&= P(o\mid s, a_{>t}, h_{\leq t}) P(s \mid a_{>t}, h_{\leq t})\propto \delta_{f(s)}(o) P(a_{>t} \mid s,h_{\leq t})P(s \mid h_{\leq t})=  \delta_{f(s)}(o)\delta_{g(s)}(a_{>t})P(s \mid h_{\leq t}).
\end{align*}
Therefore we can compute the expectation, for any value of $a_{>t}$:
\begin{align*}
    &\:\E_{P(s,o \mid a_{>t}, h_{\leq t})}\left[\log P(o \mid s, a_{>t}, h_{\leq t})- \log P(a_{>t} \mid s,o, h_{\leq t})\right] = \E_{\delta_{f(s)}(o)\delta_{g(s)}(a_{>t})P(s \mid h_{\leq t})}\left[\log \delta_{f(s)}(o) - \log \delta_{g(s)}(a_{>t})\right]\\
    &= \E_{P(s\mid h_{\leq t})}\left[\log \delta_{f(s)}(f(s)) - \log \delta_{g(s)}(g(s))\right]= \E_{P(s\mid h_{\leq t})}\left[\log 1 - \log 1\right] =0.
\end{align*}
\end{proof}

By Lemma \ref{lemma: normalisation}, we conclude that agency can be expressed as a functional of the agent's predictions and preferences, known as the \emph{expected free energy}~\cite{fristonFreeEnergyPrinciple2022,barpGeometricMethodsSampling2022}
\begin{equation}
\label{eq: EFE appendix} \tag{EFE}
    -\log P(a_{>t} \mid h_{\leq t}) = \E_{P(s,o\mid a_{>t} ,h_{\leq t})}[ \log P(s \mid a_{>t}, h_{\leq t})- \log P(s,o  \mid h_{\leq t})]
\end{equation}
Furthermore, we can rewrite the expected free energy~\eqref{eq: EFE appendix} in the following ways~\cite{barpGeometricMethodsSampling2022,sajidActiveInferenceBayesian2021}
\begin{equation*}
\label{eq: decomposition precise agent EFE}
\begin{split}
    -\log P(a_{>t} \mid h_{\leq t}) &= \kl\big[\,P(s \mid  a_{>t}, h_{\leq t}) \mid P( s\mid h_{\leq t})\,\big] +
    \E_{P(s,o \mid a_{>t}, h_{\leq t})}\big[-\log P(o \mid s , h_{\leq t})\big]\\
    &= -\mathbb{E}_{P(o \mid a , h_{\leq t})}\big[\log P\left(o \mid h_{\leq t}\right)\big]
   -\mathbb{E}_{P(o \mid a_{>t}, h_{\leq t})}\big[\kl\left[P\left(s \mid o, a_{>t}, h_{\leq t}\right) \mid P\left(s \mid a_{>t}, h_{\leq t}\right)\right]\big]\\
   &+\mathbb{E}_{P(o \mid a_{>t}, h_{\leq t})}\big[\kl\left[P\left(s \mid o,a_{>t}, h_{\leq t}\right) \mid P\left(s \mid o, h_{\leq t}\right)\right]\big]\\
   &\geq -\mathbb{E}_{P(o \mid a , h_{\leq t})}\big[\log P\left(o \mid h_{\leq t}\right)\big]
   -\mathbb{E}_{P(o \mid a_{>t}, h_{\leq t})}\big[\kl\left[P\left(s \mid o, a_{>t}, h_{\leq t}\right) \mid P\left(s \mid a_{>t}, h_{\leq t}\right)\right]\big].
\end{split}
\end{equation*}

\end{document}